%% file: main.tex
\documentclass[10pt,twocolumn,letterpaper]{article}
\usepackage[pagenumbers]{cvpr} 
\usepackage{algorithm}
\usepackage{algpseudocode}
\usepackage{multirow} 
\usepackage{bbding} 
\usepackage{silence}
\makeatletter
\robustify\@latex@warning@no@line
\makeatother
\usepackage{authblk}

\input{preamble}

\definecolor{cvprblue}{rgb}{0.21,0.49,0.74}
\usepackage[pagebackref,breaklinks,colorlinks,citecolor=cvprblue]{hyperref}

\def\paperID{4278} 
\def\confName{CVPR}
\def\confYear{2024}

\title{Diffusion-based Blind Text Image Super-Resolution}

\author[1]{Yuzhe Zhang}
\author[2]{Jiawei Zhang}
\author[2]{Hao Li}
\author[3]{Zhouxia Wang}
\author[2]{Luwei Hou}
\author[2]{\\ Dongqing Zou}
\author[1*]{Liheng Bian}

\affil[ ]{\small{$^1$Beijing Institute of Technology, $^2$SenseTime Research, $^3$The University of Hong Kong}}

\begin{document}

\twocolumn[{%
\renewcommand\twocolumn[1][]{#1}%
\maketitle
\begin{figure}[H]
\vspace{-1cm}
\hsize=\textwidth
    \centering
    \hspace{-0.4cm}
    \includegraphics[width=7 in]{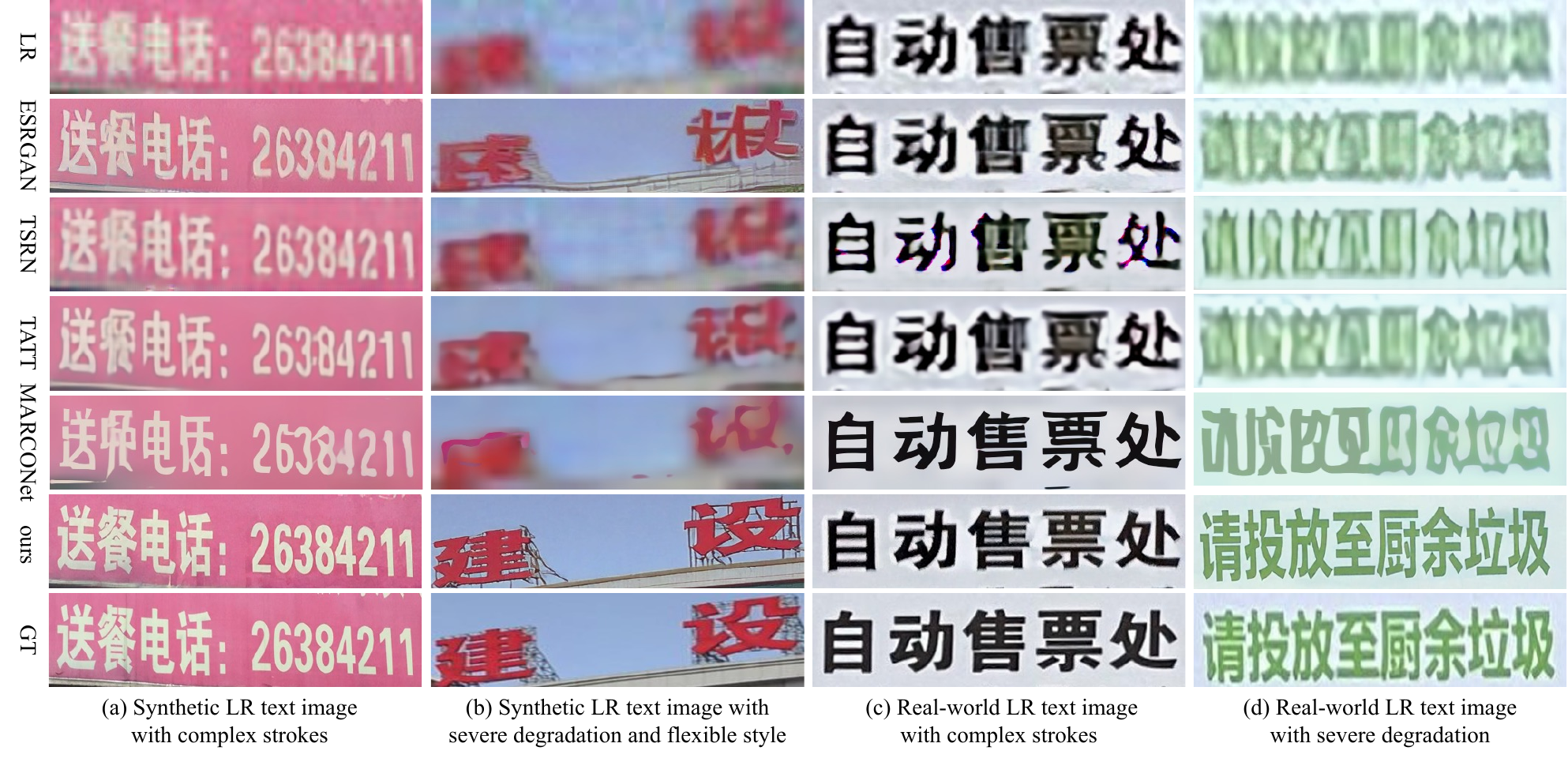}
    \caption{
    Blind text image super-resolution results between different methods on synthetic and real-world text images.
    Our method can restore text images with high text fidelity and style realness under complex strokes, severe degradation, and various text styles.
    }
    \label{fig:fig1 various degradation images} 
\end{figure}
}]

\input{sec/0_abstract}    
\input{sec/1_intro}
\input{sec/2_realted_work}

\input{sec/3_method}

\input{sec/4_experiments}

\input{sec/5_conclusion}

{
    \small
    \bibliographystyle{ieeenat_fullname}
    \bibliography{main}
}


\end{document}

%% file: preamble.tex
\usepackage[dvipsnames]{xcolor}


%% file: sec/0_abstract.tex
\begin{abstract}
\vspace{-0.2cm}
Recovering degraded low-resolution text images is challenging, especially for Chinese text images with complex strokes and severe degradation in real-world scenarios.
Ensuring both text fidelity and style realness is crucial for high-quality text image super-resolution.
Recently, diffusion models have achieved great success in natural image synthesis and restoration due to their powerful data distribution modeling abilities and data generation capabilities.
In this work, we propose an Image Diffusion Model (IDM) to restore text images with realistic styles.
For diffusion models, they are not only suitable for modeling realistic image distribution but also appropriate for learning text distribution.
Since text prior is important to guarantee the correctness of the restored text structure according to existing arts, we also propose a Text Diffusion Model (TDM) for text recognition which can guide IDM to generate text images with correct structures.
We further propose a Mixture of  Multi-modality module (MoM) to make these two diffusion models cooperate with each other in all the diffusion steps.
Extensive experiments on synthetic and real-world datasets demonstrate that our Diffusion-based Blind Text Image Super-Resolution (DiffTSR) can restore text images with more accurate text structures as well as more realistic appearances simultaneously.

\end{abstract}

%% file: sec/1_intro.tex
\section{Introduction}

Blind text image super-resolution (SR) focuses on recovering high-resolution (HR) images from low-resolution (LR) ones corrupted by various unknown degradations.
Unlike natural image super-resolution tasks which pay more attention to enriching and enhancing the image details, text fidelity and style realness should also be guaranteed in the restored text images.
Mistakenly estimated text structures, such as distorted, missing, additional, or overlapping strokes, will lead to inaccurate character semantics which is unacceptable in the restored text images.
Similarly, incorrectly generated text styles, such as changes in fonts, glyphs, colors, and poses, will make the restored text images visually unpleasant and unreal.

In order to reconstruct text images with the correct structures, existing methods \cite{ma2023text, ma2022text, chen2021scene, wang2020scene, peyrard2015icdar2015, ma2023benchmark} introduce to utilize low-level and high-level text priors to guide the restoration process by considering text structure-related losses or incorporating additional text recognition modules.
Although these methods enhance the visual appearance of characters in reconstructed images, they are difficult to restore accurate text structure when encountering text images with complex strokes or severe degradations.
To alleviate the above issues, MARCONet~\cite{li2023learning} employs a codebook to store the discrete code of each character which can be used to generate high-resolution structural details with high text fidelity.
In addition, StyleGAN~\cite{karras2020analyzing} is exploited in MARCONet to generate visually pleasant text styles.
Even though MARCONet can handle complex strokes and severe degradation to a certain extent, the predefined font styles during training limit its ability to deal with unseen and diverse text styles in the real world, leading to the unrealness and infidelity in some restored images.

Recently, diffusion models \cite{sohl2015deep, ho2020denoising} have exhibited great success in natural image synthesis \cite{nichol2021glide, song2020score, dhariwal2021diffusion, rombach2022high} and restoration \cite{kawar2022denoising, saharia2022image, luo2023refusion, fei2023generative} due to their powerful data distribution modeling and data generation capabilities. 
In this paper, we argue that diffusion model should also be suitable to model diverse text styles, which include fonts, glyphs, colors, and poses, to restore visually more pleasant and realistic text images.
As a result, we propose an Image Diffusion Model (IDM) based on stable diffusion \cite{rombach2022high} to effectively model the text styles.
To keep the text character fidelity,  IDM is conditioned on the input low-resolution image and text prediction priors.
However, accurately recognizing text from severely degraded images is challenging, and inaccurate text recognition will lead to incorrect text structures in the restoration results.
According to the analysis of \cite{hoogeboom2021argmax}, diffusion model is also appropriate to model the discrete variable distribution like text.
On this basis, we introduce to use a Text Diffusion Model (TDM) to correctly recognize texts conditioned on low-resolution input and provide text prior to help IDM restore text images with high fidelity.
It is worth emphasizing that TDM can benefit IDM and vice versa.
Therefore, we further propose a Mixture of Multi-modality module (MoM) so that these two diffusion models can cooperate with each other in all the diffusion steps.
Extensive experiments demonstrate that our Diffusion-based Blind Text Image Super-Resolution (DiffTSR) can restore text images, especially for Chinese text images with complex strokes, from degraded ones with satisfactory text fidelity and style realness simultaneously.
In summary, our work has the following main contributions:

\begin{itemize}
  \item[\textbullet] We propose to use IDM and TDM to model text image distribution and text distribution in order to restore text images with high text fidelity and style realness.
  \item[\textbullet] We propose a MoM module to make IDM and TDM closely cooperate with each other in all the diffusion steps.
  \item[\textbullet] Extensive experiments demonstrate that the proposed DiffTSR performs better than existing methods on both synthetic and real-world datasets.
\end{itemize}

%% file: sec/2_realted_work.tex
\section{Related Work}

\noindent \textbf{Blind Image Super-Resolution.} 
Blind image super-resolution (SR) aims to enhance the resolution and quality of images with complex unknown degradation in real-world scenarios. 
Recent works have made efforts to achieve more effective blind SR from two aspects: degradation model estimation~\cite{bell2019blind,gu2019blind,huang2020unfolding,maeda2020unpaired} and real-world data synthesis~\cite{fritsche2019frequency,cai2019toward,ji2020real,wei2020component}. 
The former learns the degradation model from low-resolution (LR) images in an unsupervised manner~\cite{wang2021unsupervised} and then applies non-blind SR methods. 
The latter involves synthesizing LR-HR image training pairs through a complex degradation strategy that imitates real-world degradation. 
Specifically, BSRGAN~\cite{zhang2021designing} uses a random shuffling strategy to achieve more generalized degradation data synthesis, while Real-ESRGAN~\cite{wang2021real} further enhances the complexity of image degradation through a high-order degradation modeling process. 
Although the above methods have achieved great success in blind SR of natural images, we observe that it is insufficient to effectively enhance the quality of text images without considering the specific character structures and text style.

\begin{figure*}[!t]
    \centering
    \includegraphics[width=6.8 in]{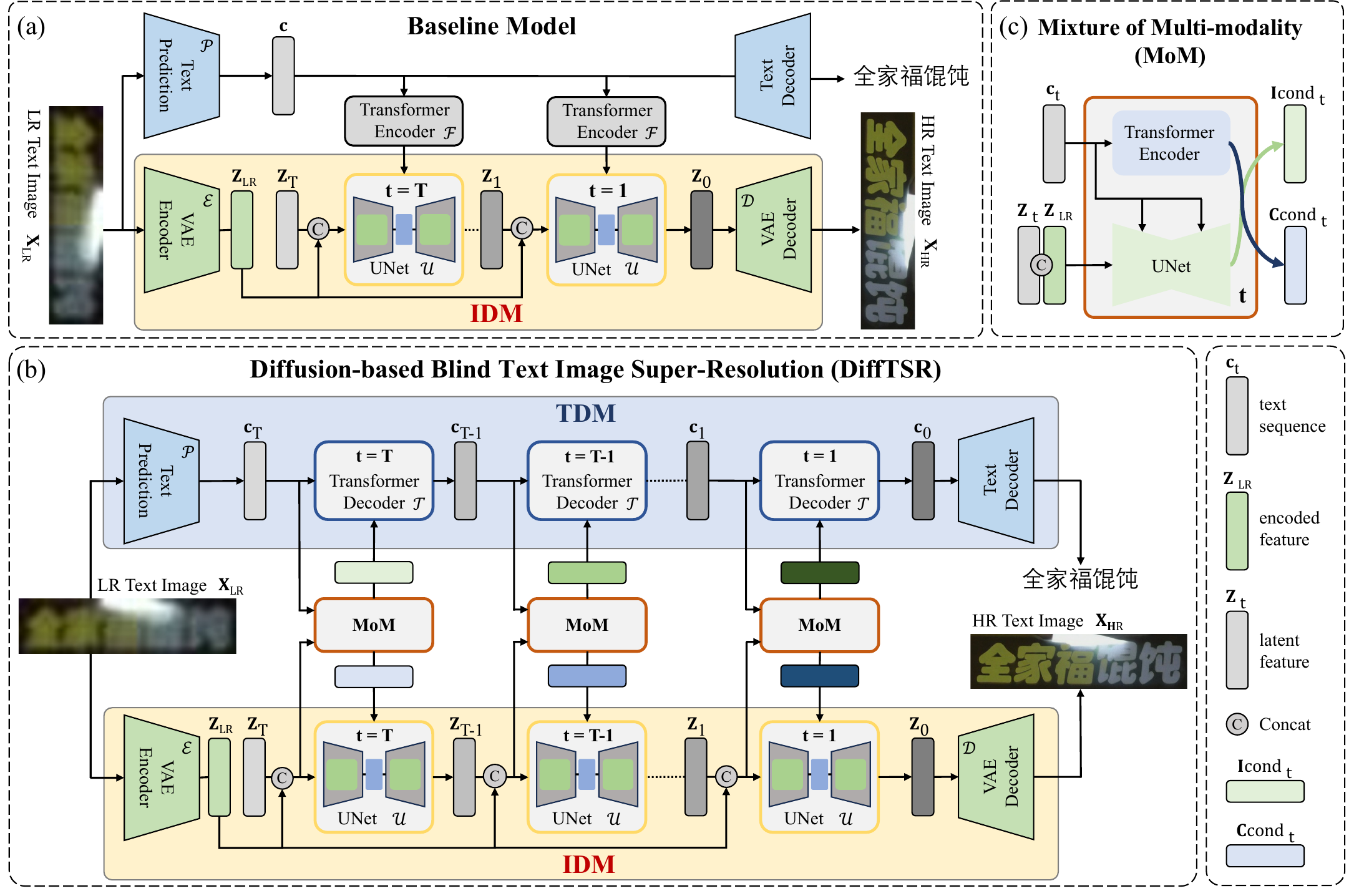}
    \caption{Overview of Diffusion-based Blind Text Image Super-Resolution (DiffTSR) along with the baseline.
    (a) Our baseline model.
    It contains an Image Diffusion Model (IDM) and a text recognition model.
    The IDM performs the diffusion-based text image super-resolution conditioned on the latent feature $\mathbf{Z}_{LR}$ from the LR image and text prior $\mathbf{c}$ which is extracted by the text recognition model from the LR image.
    (b) DiffTSR architecture. It mainly consists of three parts:
    i) IDM performs the image diffusion conditioned on $\mathbf{Z}_{LR}$ and $\mathbf{C}cond_t$ to achieve the high-realness image generation,
    ii) TDM conducts the text diffusion conditioned on $\mathbf{I}cond_t$, which starts the reverse process from the initial text prior $\mathbf{c}_T$, to achieve more accurate text prior prediction and correction,
    iii) MoM module fuses and encodes the intermediate features of IDM and TDM at the previous step, and outputs the conditions $\mathbf{C}cond_t$ and $\mathbf{I}cond_t$ for the current time step.
    IDM and TDM cooperate with each other through MoM to finally achieve text image super-resolution with high fidelity and realness.
    (c) Details of MoM.
    It fuses $\mathbf{Z}_{LR}$, $\mathbf{Z}_{t}$, and $\mathbf{c}_t$ at step $t$, and encodes them into $\mathbf{I}cond_t$ and $\mathbf{C}cond_t$ for TDM and IDM respectively.}
    \label{fig:TIR-DM model}
\end{figure*}

\noindent \textbf{Text Image Super-Resolution.} 
Text image super-resolution aims to enhance the details of the image meanwhile improving the readability of the text, \ie the accuracy of text recognition. 
Recent research mainly focuses on exploring the guidance of text recognition priors and character structure priors to improve the performance of text image SR.
Specifically, based on the characteristics of text images, existing works mainly exploit text-related prior information from three aspects to constrain text image super-resolution: text aware loss~\cite{wang2019textsr,wang2020scene,chen2021scene}, text recognition prior~\cite{xu2017learning, zhao2022c3, ma2022text, mou2020plugnet}, and text structure prior~\cite{li2023learning}.
Previous research demonstrates that the text priors play an important role in text structure enhancement.
However, most of them do not fully utilize text prior information and cannot restore text images with diverse text styles, severe degradation, or complex strokes.

\noindent \textbf{Diffusion Model.}
Diffusion model~\cite{ho2020denoising} has attracted great attention, due to its impressive performance in image synthesis~\cite{dhariwal2021diffusion,rombach2022high,gu2022vector}, and controllable image generation~\cite{ho2022imagen,saharia2022photorealistic,ruiz2023dreambooth,zhang2023adding,mou2023t2i}.
Benefiting from the powerful data distribution modeling capability of diffusion models, recent research~\cite{saharia2022image,li2022srdiff,kawar2022denoising,wang2022zero,wang2023exploiting} also achieves impressive performance in image super-resolution by utilizing the diffusion prior.
In addition, existing research has shown that diffusion models are also suitable for modeling discrete data~\cite{hoogeboom2021argmax}, such as text~\cite{gong2022diffuseq}, segmentation map~\cite{zbinden2023stochastic}, \etc.
In this work, we aim to explore the collaboration between image diffusion models and text diffusion models, and achieve high-quality text image super-resolution with high text fidelity and style realness.

%% file: sec/3_method.tex
\section{Methodology}
\subsection{Overview}

In this paper, we propose to use the diffusion model to restore degraded text images by considering text prior.
We first propose a baseline model which is shown in Figure~\ref{fig:TIR-DM model}~(a) and described in Section~\ref{sec:IDM}.
It uses a text recognition model to provide text prior.
Then, the proposed Image Diffusion Model (IDM) is used to restore the text images conditioned on the text prior.
Even though the above baseline model can restore degraded text images with relatively acceptable fidelity, it will produce distorted text structures when encountered with severe degradation.
Besides IDM, the proposed Diffusion-based Blind Text Image Super-Resolution (DiffTSR) also contains a Text Diffusion Model (TDM) and a Mixture of Multi-modality module (MoM) based on the assumption that more accurate text recognition information can be beneficial for IDM to generate a more realistic image; meanwhile, a higher-quality text image can benefit for better recognition.
In DiffTSR, TDM is a diffusion model that gradually recognizes text sequence with given image information. 
As to MoM, it is like a bridge to connect IDM with TDM.
It provides updated text prior to IDM and image information to TDM during the diffusion process.
In this way, the proposed DiffTSR can restore text images with high style realness and text fidelity simultaneously.
The overall architecture of DiffTSR is shown in Figure~\ref{fig:TIR-DM model}~(b) and described in Section~\ref{sec:TIR-DM}.

\subsection{Baseline} \label{sec:IDM}
Our baseline model is illustrated in Figure~\ref{fig:TIR-DM model}~(a). It mainly consists of two parts, 1) an Image Diffusion Model (IDM), 2) a text recognition model. 
The text recognition model estimates text sequence $\mathbf{c}$ from the low-resolution text images $\mathbf{X}_{LR}$ as text prior in every diffusion step, and IDM implements the image super-resolution through the diffusion reverse process conditioned on $\mathbf{c}$ and $\mathbf{X}_{LR}$.

To model the distribution of real-world text images and achieve realistic image generation, the proposed IDM is based on Stable Diffusion~\cite{rombach2022high}.
IDM performs the diffusion forward and reverse process in the latent space through a VAE encoder $\mathcal{E}$ and decoder $\mathcal{D}$.
After encoding HR image $\mathbf{X}$ into latent space by $\mathcal{E}$ as $\mathbf{Z}=\mathcal{E}(\mathbf{X})$, IDM sequentially adds noises into  $\mathbf{Z}$ at time step $t$ as $\mathbf{Z}_t$ and a sequence of noise prediction network $\mathcal{U}_{\theta}$ is used to gradually remove the noises in the reverse process.
In order to make the restoration results consistent with the input LR image, encoded feature $\mathbf{Z}_{LR}=\mathcal{E}(\mathbf{X}_{LR})$ is considered as a condition in $\mathcal{U}_{\theta}$ by concatenation with $\mathbf{Z}_{t}$.
Meanwhile, text prior is also considered as another condition in $\mathcal{U}_{\theta}$.
Specifically, we encode $\mathbf{c}$, which is estimated from the text recognition model $\mathcal{P}$ in \cite{chen2021scene}, through a transformer encoder $\mathcal{F}_\psi$, and fuse the encoded feature $\mathcal{F}_\psi(\mathbf{c})$ into the intermediate layers of $\mathcal{U}_{\theta}$ by the cross-attention mechanism.

The details of the sampling process of our baseline model can be described as follows.
We first utilize the text recognition model $\mathcal{P}$ to predict the text sequence $\mathbf{c}$ from the LR text image $\mathbf{X}_{LR}$.
After that, IDM starts the reverse process and repeats the denoising step $\mathcal{U}_{\theta}$ conditioned on the latent feature $\mathbf{Z}_{LR}$ extracted from LR image $\mathbf{X}_{LR}$ by VAE encoder $\mathcal{E}$ and text prior  $\mathcal{F}_\psi(\mathbf{c})$ extracted from $\mathbf{c}$ by a transformer encoder  $\mathcal{F}_\psi$ until obtaining $\mathbf{Z}_0$.
Then the restored text image can be reconstructed through VAE decoder as $\mathcal{D}(\mathbf{Z}_0)$.
Our baseline model can restore text images with high realness, which benefits from the ability of IDM to generate realistic details.

\begin{figure}[!t]
    \centering
    \setlength{\abovecaptionskip}{0.1cm}
    \hspace{-0.5cm}
    \includegraphics[width=\linewidth]{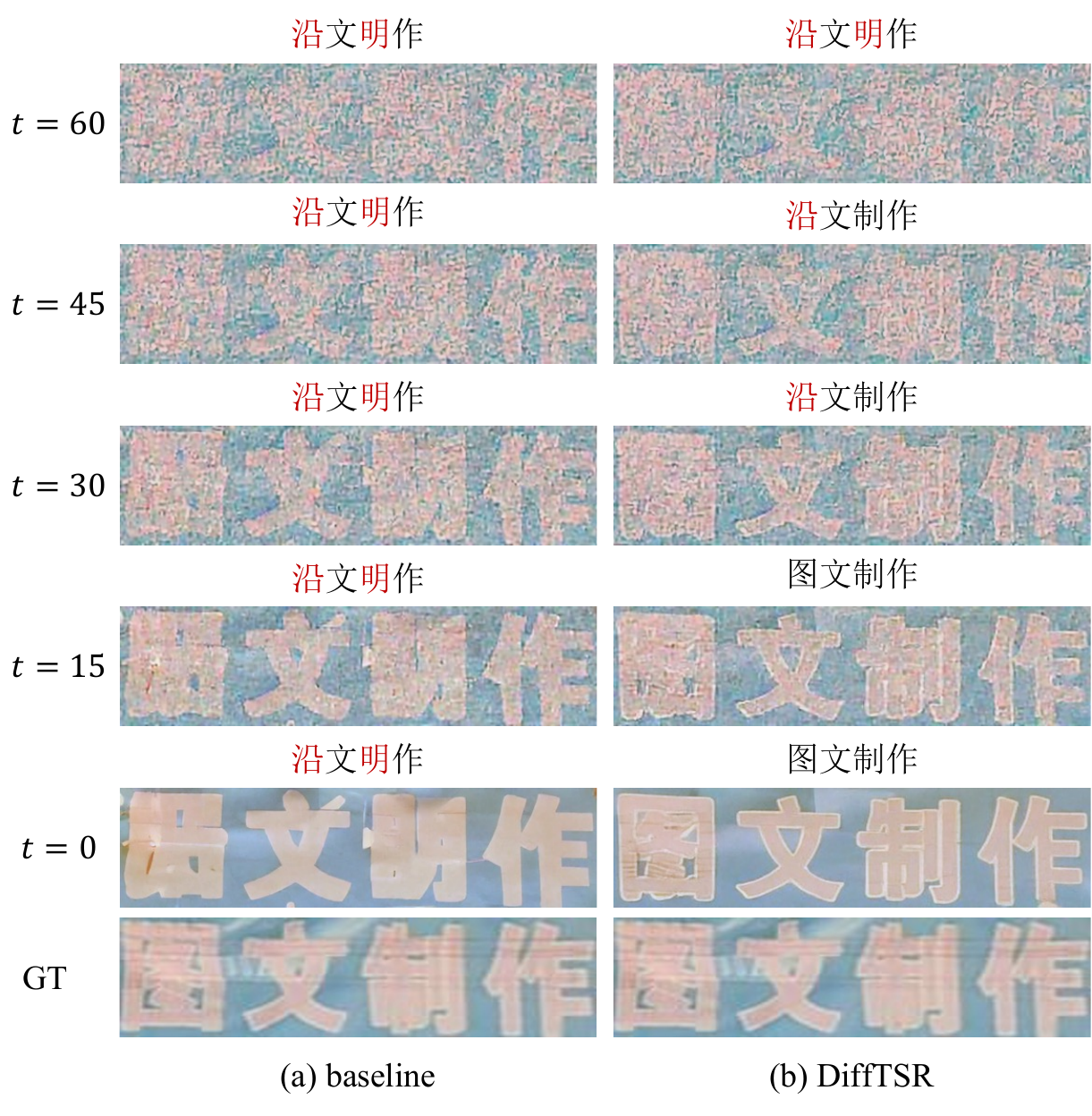}
    \caption{Motivation.
    To provide text prior for text image restoration, the baseline model recognizes text from degraded images which is inaccurate when the degradation is severe.
    With inaccurate text prior, the baseline model cannot restore text image with high text fidelity which is shown in (a).
    The proposed TDM and IDM can benefit from each other through MoM in DiffTSR and gradually recognizes more accurate text sequence and restore higher-quality text image through the reverse diffusion process which is shown in (b).
    The text sequences above each super-resolution result at different time steps are the recognized text characters used for blind image super-resolution and the characters in \textcolor{red}{red} are the mistakenly estimated ones.
    }
    \label{fig:motivation}
\end{figure}

\subsection{Diffusion-based Blind Text Image Super-Resolution} \label{sec:TIR-DM}
Even though the baseline model above can effectively restore low-resolution text images, it will still generate unpleasant results when encountered with severe degradation which is shown in Figure~\ref{fig:motivation}~(a).
This is because the text recognition model cannot work well with highly distorted text images and IDM cannot restore text images with high text fidelity under inaccurate text prior.
In this subsection, we propose Diffusion-based Blind Text Image Super-Resolution (DiffTSR) to restore text images with high text fidelity and style realness by jointly optimizing image restoration and text recognition in every diffusion step.
Besides IDM the same as the baseline model, DiffTSR also contains a Text Diffusion Model (TDM) and a Mixture of Multi-modality module (MoM).
The details of TDM, MoM, and the whole DiffTSR are described as follows.

Existing works \cite{hoogeboom2021argmax, fujitake2023diffusionstr} indicate that the diffusion model can not only model image distribution but also model discrete data such as text.
To model the distribution of text sequence $\mathbf{c} $, TDM also follows the Markov chain of the diffusion process that slowly adds random noises to the text sequence in the forward process and then learns the reverse process to reconstruct the text sequence from the noisy data.
Unlike IDM which is a continuous diffusion model and the added noises satisfy Gaussian distribution, TDM is a discrete one.
Similar to \cite{hoogeboom2021argmax, zbinden2023stochastic}, TDM assumes the transition distribution $q\left(\boldsymbol{c}_t\mid\boldsymbol{c}_{t-1}\right)$ follows a categorical distribution in the forward process.
With this assumption, TDM proposes to use a Transformer Decoder $\mathcal{T}_\eta$ to remove noises from $\mathbf{c}_{t}$ and generate $\mathbf{c}_{pred}$.
To make the text sequence modeling more consistent with the context of the input image,  $\mathbf{I}cond_t$, which contains image information estimated by MoM, is mapped to the intermediate layer of $\mathcal{T}_\eta$ through the cross-attention mechanism.
TDM benefits from text modeling capability as well as image conditions guidance and produces more reasonable and accurate text sequences consistent with the LR image.

As text recognition can benefit text image super-resolution and vice versa, we propose a Mixture of Multi-modality module (MoM) for joint optimization as shown in Figure~\ref{fig:TIR-DM model}~(c).
MoM consists of two time-aware modules, a UNet and a Transformer encoder.
The UNet of MoM at time step $t$ first extracts image information from the concatenated $\mathbf{Z}_t$ and $\mathbf{Z}_{LR}$.
Then, $\mathbf{c}_t$ is mapped into the intermediate layer of  UNet through cross-attention mechanism.
UNet fuses and encodes the multi-modality information into the image condition $\mathbf{I}cond_t$ for TDM at each time step, thereby adaptively generating image conditions that are more suitable for TDM to achieve higher recognition accuracy.
At the same time, the Transformer encoder of MoM receives corrected characters $\mathbf{c}_t$ from the previous step of TDM, and encodes $\mathbf{c}_t$ into the characters embedding space of IDM as the text condition $\mathbf{C}cond_t$.
In summary, 
\begin{equation}\label{eq:MoM_fusion}
\begin{aligned}   
\left[{\mathbf{I}cond}_t, {\mathbf{C}cond}_t\right] = MoM_\phi([\mathbf{Z}_{LR}, \mathbf{Z}_t], \mathbf{c}_t, t)\text{,}
\end{aligned}
\end{equation}
where $\mathbf{C}cond_t$ and $\mathbf{I}cond_t$ serve as the condition for IDM and TDM at time step $t$, respectively.

After introducing IDM, TDM and MoM,  the sampling process of DiffTSR is shown in Algorithm~\ref{Alg:TIR-DM Sampling} and Figure~\ref{fig:TIR-DM model}~(b).
To begin with, we extract features $\mathbf{Z}_{LR}$ from LR image $\mathbf{X}_{LR}$ through VAE encoder $\mathcal{E}$.
At the same time, we randomly sample $\mathbf{Z}_T$ with Gaussian distribution, and get the initially estimated text $\mathbf{c}_T=\mathcal{P}(\mathbf{X}_{LR})$ from the LR image $\mathbf{X}_{LR}$.
Note that TDM starts the reverse process from the initial estimated text rather than the random sampled ones.
After the initialization process, IDM uses the UNet $\mathcal{U}$ to remove noises from the latent features with given $\mathbf{Z}_{LR}$, the denoised feature from the previous step $\mathbf{Z}_{t}$ as well as text prior $\mathbf{C}cond_t$ from MoM at every time step.
At the same time, TDM uses Transformer Decoder $\mathcal{T}$ to estimate the latent state of the text sequence with given the state $\mathbf{c}_t$ from the previous time step and the image condition $\mathbf{I}cond_t$ from MoM.
With $T$ collaborative diffusion steps, $\mathbf{Z}_0$ can be estimated and VAE Decoder $\mathcal{D}$ is used to reconstruct HR text image $\mathbf{X}_0$ with high fidelity and realness.
Thanks to the joint optimization strategy through MoM, the proposed DiffTSR can gradually restore HR text image by IDM and more accurate text sequence in TDM which is shown in Figure~\ref{fig:motivation}~(b).
For more details about the sampling and training strategy of DiffTSR, please see the supplementary material.

\algnewcommand{\NoNumber}[1]{\Statex \hspace{0em} \(\triangleright\) #1}
\begin{algorithm}[!t]
\caption{DiffTSR Sampling}
\label{Alg:TIR-DM Sampling}
\begin{algorithmic}[1]
\NoNumber \textbf{input :}  LR Text Image $\mathbf{X}_{LR}$ 
\NoNumber \textbf{output :} HR Text Image $\mathbf{X}_{HR}$
\State $\mathbf{Z}_{LR}=\mathcal{E}(\mathbf{X}_{LR})$
\State $\mathbf{Z}_T\sim \mathcal{N}(0, I)$
\State $\mathbf{c}_T=\mathcal{P}(\mathbf{X}_{LR})$
\For{$t = T, \dots, 1$}
\State $\mathbf{z} \sim \mathcal{N}(\mathbf{0}, \mathbf{I})$ if $t>1$, else $\mathbf{z}=\mathbf{0}$
\State $[{\mathbf{I}cond}_t, {\mathbf{C}cond}_t] = MoM_\phi([\mathbf{Z}_{LQ}, \mathbf{Z}_t], \mathbf{c}_t, t)$ 
\State $\boldsymbol{\epsilon}_{pred,t}=\mathcal{U}_{\theta}([\mathbf{Z}_t, \mathbf{Z}_{LR}], \mathbf{C}cond_t, t)$
\Statex \hspace{0.5em} \textcolor{gray}{// IDM sampling based on stable diffusion \cite{rombach2022high}}
\State $\mathbf{Z}_{t-1}=\frac{1}{\sqrt{\alpha_t^{IDM}}}\left(\mathbf{Z}_t-\frac{1-\alpha_t^{IDM}}{\sqrt{1-\bar{\alpha}_t^{IDM}}} \epsilon_{pred,t}\right)+\sigma_t \mathbf{z}$
\State $\mathbf{c}_{pred,t}=\mathcal{T}_\eta(\mathbf{c}_t, \mathbf{I}cond_t, t)$
\State $\tilde{\boldsymbol{\pi}}= \small
{ \left[\alpha_t^{TDM} \mathbf{c}_t\!+\!\frac{1\!-\!\alpha_t^{TDM}}{K}\right] \! \odot \! \left[\bar{\alpha}_{t-1}^{TDM} \mathbf{c}_{pred, t}\!+\!\frac{1\!-\!\bar{\alpha}_{t-1}^{TDM}}{K}\right]}$
\State $\boldsymbol{\pi}_{\text {post }}\left(\mathbf{c}_t, \mathbf{c}_{pred, t}\right)\!=\!\frac{\tilde{\boldsymbol{\pi}}}{\sum_{k=1}^K \tilde{\pi}_k }$
\Statex \hspace{0.5em} \textcolor{gray}{// TDM sampling based on multinomial diffusion \cite{hoogeboom2021argmax}}
\State $\mathbf{c}_{t-1} \sim \mathcal{C}\left(\mathbf{c}_{t-1} \mid \boldsymbol{\pi}_{\text {post }}\left(\mathbf{c}_t, \mathbf{c}_{pred,t}\right)\right)$ if $t>1$ else $\mathbf{c}_{t-1} \sim\mathcal{C}\left(\mathbf{c}_0\mid \mathbf{c}_{pred,t}\right)$
\EndFor
\State $\mathbf{X}_0=\mathcal{D}(\mathbf{Z}_0)$
\State \textbf{return }$\mathbf{X}_0$
\Statex \textcolor{gray}{// $\mathcal{C}$ denotes the categorical distribution with probability parameters after $\mid$. }
\Statex \textcolor{gray}{// $1-{\alpha}_{t-1}^{IDM}$ and $1-{\alpha}_{t-1}^{TDM}$ are the noise schedule for IDM and TDM.}
\Statex \textcolor{gray}{// The processing details from Ln. 9 to Ln. 11 are described in \cite{hoogeboom2021argmax}.}
\end{algorithmic}
\end{algorithm}

%% file: sec/4_experiments.tex
\section{Experiments}

\begin{table*}[htbp]
  \small
  \renewcommand{\arraystretch}{0.8}
  \centering
   \setlength{\tabcolsep}{2.8mm}{
    \begin{tabular}{l|ccccc|ccccc}
    \toprule
    \multirow{2}[0]{*}{method} & \multicolumn{5}{c|}{$\times$ 2}               & \multicolumn{5}{c}{$\times$ 4} \\
          & PSNR ↑  & LPIPS ↓ & FID ↓ & ACC ↑ & NED ↑ & PSNR ↑  & LPIPS ↓ & FID ↓ & ACC ↑ & NED ↑ \\
    \midrule
    SRCNN & 23.73 & 0.338 & 54.47 & 0.7856 & 0.7991 & 20.74 & 0.501 & 116.5 & 0.6031 & 0.6160 \\
    ESRGAN & 24.75 & 0.191 & 9.308 & 0.8112 & 0.8239 & 20.90  & 0.310  & 21.86 & 0.6179 & 0.6272 \\
    NAFNet & 25.04 & 0.286 & 37.42 & 0.8083 & 0.8212 & 21.82 & 0.447 & 87.93 & 0.6451 & 0.6573 \\
    TSRN  & 20.86 & 0.392 & 70.75 & 0.7805 & 0.7937 & 19.41 & 0.535 & 137.3 & 0.6149 & 0.6267 \\
    TBSRN & 24.43 & 0.282 & 57.61 & 0.8018 & 0.8156 & 21.56 & 0.442 & 132.6 & 0.6360  & 0.6486 \\
    TATT  & 24.87 & 0.291 & 58.73 & 0.7911 & 0.8041 & 21.84 & 0.453 & 107.6 & 0.6273 & 0.6403 \\
    MARCONet & 20.77 & 0.374 & 94.60  & 0.6934 & 0.7068 & 19.33 & 0.436 & 108.5 & 0.5123 & 0.5241 \\
    ours & \textbf{25.08} & \textbf{0.156} & \textbf{5.906} & \textbf{0.8594} & \textbf{0.8718} & \textbf{21.85} & \textbf{0.231} & \textbf{8.482} & \textbf{0.8350} & \textbf{0.8471} \\
    \bottomrule
    \bottomrule
    \end{tabular}}%
  \caption{Quantitative comparison for the synthetic dataset CTR-TSR-Test with different methods including SRCNN~\cite{dong2015image}, ESRGAN~\cite{wang2018esrgan},  NAFNet~\cite{chen2022simple}, TSRN~\cite{wang2020scene}, TBSRN~\cite{chen2021scene}, TATT~\cite{ma2022text}, MARCONet~\cite{li2023learning} and our method for $\times 2$ and $\times 4$ blind text image super-resolution.}
  \label{tab:fudanvi dataset eval result}%
\end{table*}%

\begin{table*}[htbp]
  \small
  \renewcommand{\arraystretch}{0.8}
  \centering
    \setlength{\tabcolsep}{2.8mm}{
    \begin{tabular}{l|ccccc|ccccc}
    \toprule
    \multirow{2}[0]{*}{method} & \multicolumn{5}{c|}{$\times$ 2}               & \multicolumn{5}{c}{$\times$ 4} \\
          & PSNR ↑  & LPIPS ↓ & FID ↓ & ACC ↑ & NED ↑ & PSNR ↑  & LPIPS ↓ & FID ↓ & ACC ↑ & NED ↑ \\
    \midrule
    SRCNN & 17.87 & 0.224 & 54.44 & 0.7922 & 0.8936 & 16.63 & 0.364 & 128.1 & 0.7101 & 0.8018 \\
    ESRGAN & 18.19 & 0.231 & 28.70  & 0.7929 & 0.8945 & 16.84 & 0.407 & 83.22 & 0.7121 & 0.8047 \\
    NAFNet & 17.86 & 0.216 & 50.42 & 0.7916 & 0.8925 & 16.76 & 0.359 & 118.1 & 0.7122 & 0.8023 \\
    TSRN  & 15.54 & 0.327 & 106.0   & 0.7892 & 0.8902 & 15.22 & 0.418 & 148.5 & 0.6963 & 0.7873 \\
    TBSRN & 17.34 & 0.236 & 71.29 & 0.7915 & 0.8932 & 16.51 & 0.367 & 130.8 & 0.7050  & 0.7960 \\
    TATT  & 17.76 & 0.247 & 59.62 & 0.7953 & 0.8951 & 16.79 & 0.422 & 118.3 & 0.7214 & 0.8135 \\
    MARCONet & 16.72 & 0.363 & 92.11 & 0.7743 & 0.8738 & 16.04 & 0.397 & 103.1 & 0.6638 & 0.7411 \\
   ours & \textbf{18.88} & \textbf{0.211} & \textbf{25.08} & \textbf{0.9085} & \textbf{0.9247} & \textbf{17.49} & \textbf{0.336} & \textbf{70.59} & \textbf{0.8475} & \textbf{0.8747} \\
    \bottomrule
    \bottomrule
    \end{tabular}}%
  \caption{Quantitative comparison for the real-world dataset RealCE~\cite{ma2023benchmark} with different methods including SRCNN~\cite{dong2015image}, ESRGAN~\cite{wang2018esrgan},  NAFNet~\cite{chen2022simple}, TSRN~\cite{wang2020scene}, TBSRN~\cite{chen2021scene}, TATT~\cite{ma2022text}, MARCONet~\cite{li2023learning} and our method for $\times 2$ and $\times 4$ blind text image super-resolution.}
  \label{tab:realce dataset eval result}%
\end{table*}%

\begin{figure*}[ht]
    \centering
    \hspace{-1cm}
    \includegraphics[width=6.8 in]{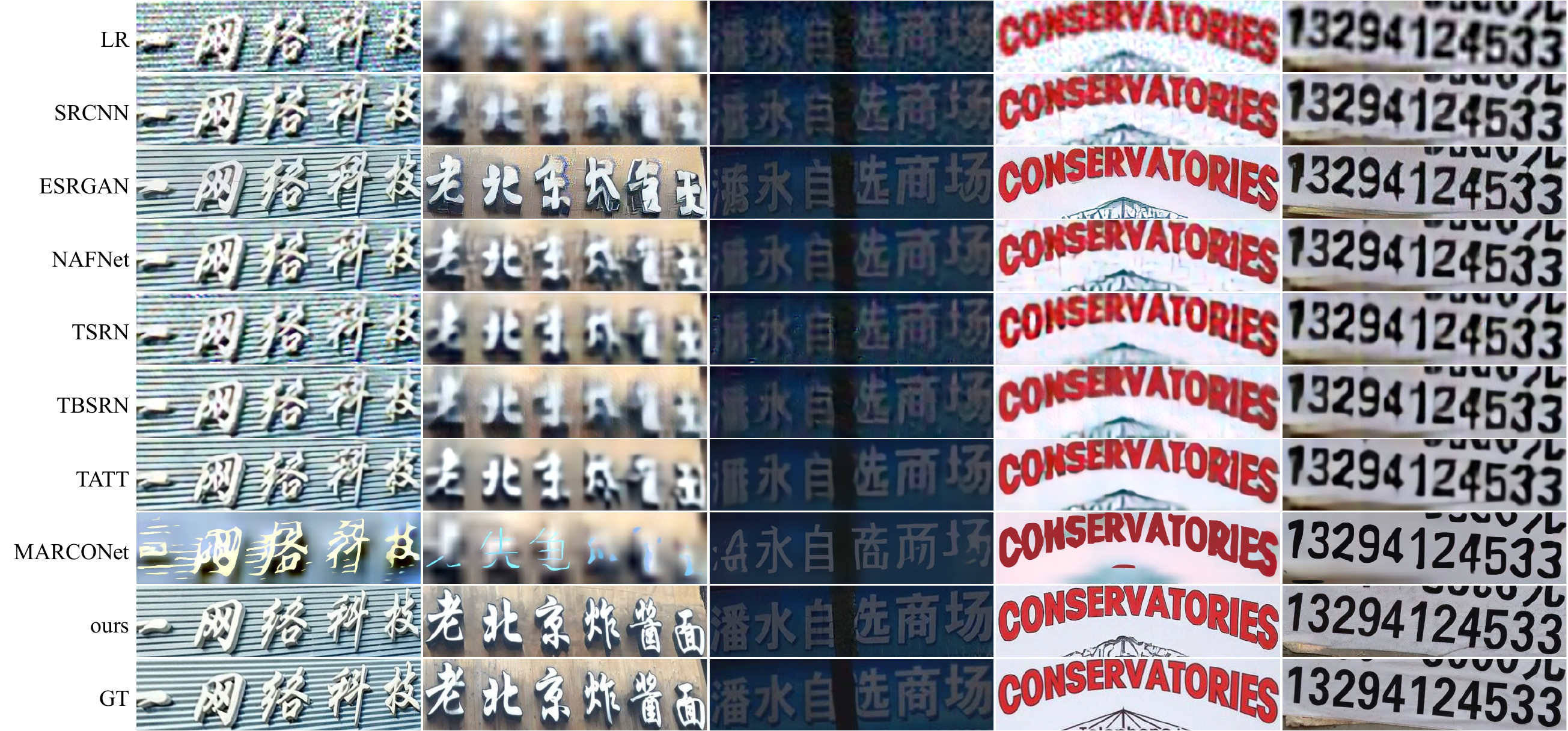}
    \caption{Qualitative comparison for the synthetic dataset CTR-TSR-Test with different methods including SRCNN~\cite{dong2015image}, ESRGAN~\cite{wang2018esrgan},  NAFNet~\cite{chen2022simple}, TSRN~\cite{wang2020scene}, TBSRN~\cite{chen2021scene}, TATT~\cite{ma2022text}, MARCONet~\cite{li2023learning} and our method for $\times 4$ super-resolution.
    }
    \label{fig:visual-comp-1}
\end{figure*}

\begin{figure*}[ht]
    \centering
    \hspace{-1cm}
    \includegraphics[width=6.8 in]{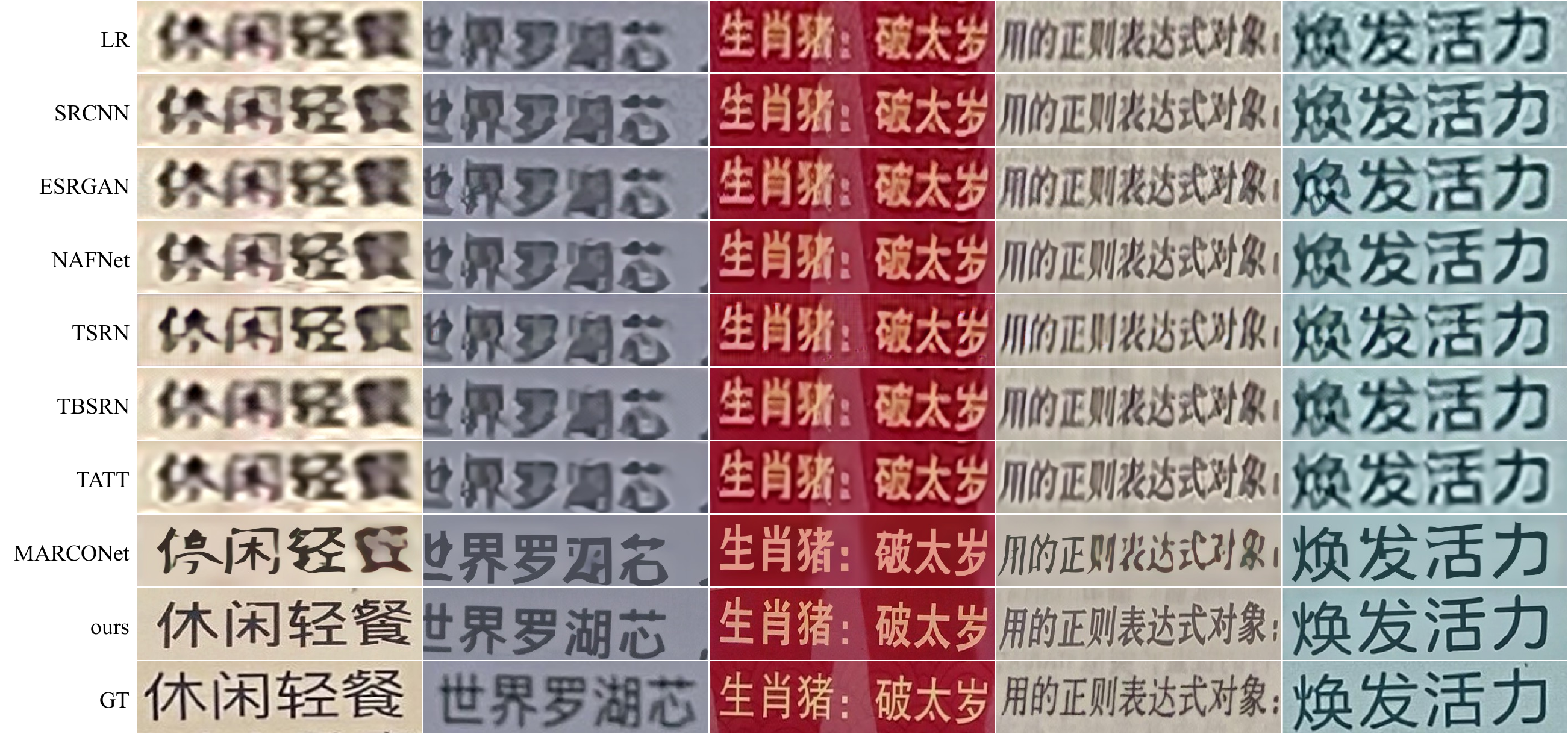}
    \caption{Qualitative comparison for the real-world dataset RealCE~\cite{ma2023benchmark} with different methods including SRCNN~\cite{dong2015image}, ESRGAN~\cite{wang2018esrgan}, NAFNet~\cite{chen2022simple}, TSRN~\cite{wang2020scene}, TBSRN~\cite{chen2021scene}, TATT~\cite{ma2022text}, MARCONet~\cite{li2023learning} and our method for $\times 4$ super-resolution.}
    \label{fig:visual-comp-2}
\end{figure*}

\begin{table*}[htbp]
  \small
  \renewcommand{\arraystretch}{0.8}
  \centering
    \setlength{\tabcolsep}{2mm}{
    \begin{tabular}{c|c|c|c|c|ccccc}
    \toprule
    \multirow{2}[4]{*}{method} & \multicolumn{4}{c|}{Settings} & \multicolumn{5}{c}{CTR-TSR-Test  / RealCE~\cite{ma2023benchmark}} \\
\cmidrule{2-10}          & IDM   & TR    & TDM   & MoM   & PSNR ↑  & LPIPS ↓ & FID ↓ & ACC ↑ & NED ↑ \\
    \midrule
    exp1   & \Checkmark     & \XSolidBrush  & \XSolidBrush     & \XSolidBrush     & 20.61 / 16.86      & 0.289 / 0.375    & 13.91 / 74.63    & 0.6342 / 0.6953    & 0.6450 / 0.7801 \\
    exp2  & \Checkmark     & \Checkmark  & \XSolidBrush     & \XSolidBrush     & 21.58 / 17.05      & 0.253 / 0.398     & 9.968 / 76.24     & 0.6811 / 0.7241     & 0.6903 / 0.8183 \\
    exp3   & \Checkmark     & \Checkmark  & \Checkmark     & \XSolidBrush     & 21.63 / 17.22      & 0.233 / 0.340      & 8.925 / 73.85     & 0.7412 / 0.7936     & 0.7530 / 0.8417 \\
    ours  & \Checkmark     & \Checkmark  & \Checkmark     & \Checkmark     & \textbf{21.85} / \textbf{17.49}     & \textbf{0.231} / \textbf{0.336}     & \textbf{8.482} / \textbf{70.59}     & \textbf{0.8350} / \textbf{0.8475}     & \textbf{0.8471} / \textbf{0.8747} \\
    \bottomrule
    \bottomrule
    \end{tabular}}
  \caption{Ablation study based on $\times 4$ super-resolution to validate the effectiveness of initial text recognition (TR), TDM, and MoM.
  For the detailed settings of different methods, please see Sec.~\ref{sec:ablation}.
  }
  \label{tab:addlabel}%
\end{table*}%

\subsection{Experimental Settings}
\noindent \textbf{Training Datasets.}
In this work, we mainly focus on blind text image super-resolution for Chinese characters in the real world. 
In order to obtain amount of HR Chinese text images along with text annotations, we use the large-scale real-world Chinese text images dataset CTR~\cite{chen2021benchmarking}.
To select the images as the ground truth in the training process, we preprocess the CTR training set by the following  steps:
i) remove the images with a resolution smaller than $64$ pixels,
ii) only retain images with a width-to-height ratio greater than $2$,
iii) only retain images with the length of text annotations not larger than 24,
iv) resize the image to $128 \times 512$.
Then, there are 63,644 HR text images $\mathbf{X}_{HR}$ remaining with text annotations $\mathbf{c}$, and we refer to this dataset as the CTR-TSR-Train.
The degradation pipeline proposed in BSRGAN~\cite{zhang2021designing} and Real-ESRGAN~\cite{wang2021real} is used to generate LR text images $\mathbf{X}_{LR}$.
\noindent \textbf{Testing Datasets.}
We evaluate our method on both synthetic and real-world datasets for $\times 2$ and $\times 4$ blind super-resolution. 
For the synthetic testing set, we use the same preprocessing and degradation strategy as in  CTR-TSR-Train to generate CTR-TSR-Test.
The images are selected from the testing set of CTR and there are 8,089 samples in total.
For real-world dataset, we use the RealCE~\cite{ma2023benchmark} testing set, which is a recently proposed real-world Chinese-English benchmark dataset.
We remove the images with more than 24 characters or images with severe LR-HR misalignment.
Finally, we obtain 1531 LR-HR pairs for real-world testing set. 

\noindent \textbf{Compared Methods and Evaluation Metrics.}
In order to validate the effectiveness of our method, we compare our DiffTIR with the natural image super-resolution methods ($i$.$e$., SRCNN~\cite{dong2015image}, ESRGAN~\cite{wang2018esrgan}, and NAFNet~\cite{chen2022simple}) and text image super-resolution methods ($i$.$e$., TSRN~\cite{wang2020scene}, TBSRN~\cite{chen2021scene}, TATT~\cite{ma2022text}, and MARCONet~\cite{li2023learning}) respectively.
For a fair comparison, we revise their implementation to handle $\times 2$ and $\times 4$ image upsampling, and finetune them with CTR-TIR-train dataset.
Moreover, we employ 5 metrics to evaluate the performance of the above methods on text image restoration.
We adopt the peak signal-to-noise ratio (PSNR) and  learned perceptual image patch similarity (LPIPS)~\cite{zhang2018unreasonable} to evaluate the distance between the restored image and reference image in the image space and feature space, respectively.
To further evaluate the realness of the restored image, we employ the Fréchet Inception Distance (FID)~\cite{heusel2017gans}. 
To better evaluate the text fidelity of the restored text image, we employ the word accuracy (ACC), and normalized edit distance (NED)~\cite{ma2023benchmark}.
Particularly, we adopt pre-trained TransOCR~\cite{chen2021scene, chen2021benchmarking} as the text recognition model for 
 ACC and NED.

\subsection{Quantitative Comparison}
We show the quantitative comparison on the synthetic test dataset CTR-TSR-Test and the real-world test dataset RealCE.
As shown in Table~\ref{tab:fudanvi dataset eval result}, DiffTSR performs better than the compared methods in all metrics. 
It achieves the best PSNR which demonstrates it can accurately reconstruct the HR images.
Benefiting from the powerful text image modeling ability, our method shows better performance in LPIPS and FID which indicates higher realness in the restored images.
Our method also performs better in terms of ACC and NED which demonstrates it can effectively keep the text fidelity with the text prior provided by TDM.
Also please note that our method still shows the best performance on RealCE without any fine-tuning on RealCE training set, as shown in Table~\ref{tab:realce dataset eval result}, indicating its strong generalization performance and powerful modeling ability for real-world text images.

\subsection{Qualitative Comparison}
The qualitative results of the synthetic dataset CTR-TSR-Test are shown in Figure~\ref{fig:visual-comp-1}.
Most of the super-resolution methods and text super-resolution ones, e.g. SRCNN~\cite{dong2015image}, NAFNet~\cite{chen2022simple}, TSRN~\cite{wang2020scene}, TBSRN~\cite{chen2021scene}, and TATT~\cite{ma2022text}, restore LR images worse than our method which demonstrates the ability of the proposed IDM to generate text images with high realness.
With the strong generation ability of GAN, ESRGAN~\cite{wang2018esrgan} can restore more realistic images (the first result).
However, it will also generate artifacts when the degradation is too severe (the second and third results).
Even though MARCONet~\cite{li2023learning} has the capability to restore more visually pleasant text structures because of the codebook, it will generate artifacts (the first result) when the text style is not considered during its training process.
In addition, the text prior of MARCONet is inaccurate when the degradation is severe (the second example) or encountered with occlusion (the third example).
These will make the restoration results of MARCONet with less text fidelity.
With the help of the proposed TDM to model the text sequence and MoM to simultaneously optimize IDM and TDM, our method can generate HR images with higher text fidelity.

We also compare different methods based on the real-world dataset RealCE~\cite{ma2023benchmark}.
Note that all the methods are not trained on the training set of RealCE to evaluate the generalization ability when encountered with unknown styles and degradation.
The results shown in Figure~\ref{fig:visual-comp-2} demonstrate that most of the methods, such as SRCNN~\cite{dong2015image}, ESRGAN~\cite{wang2018esrgan}, NAFNet~\cite{chen2022simple}, TSRN~\cite{wang2020scene}, TBSRN~\cite{chen2021scene}, and TATT~\cite{ma2022text}, can hardly remove degradation in this real-world dataset.
Although MARCONet~\cite{li2023learning} can restore HR text image to some extent, it will still generate some inaccurate and unpleasant strokes in the results.
With the strong distribution modeling abilities of IDM, the proposed methods can generalize well on real-world scenarios and restore text images with high style realness as well as text fidelity.

\begin{figure}
	 \vspace{-0.5cm}
	\centering
	\hspace{-0.5cm}
	\includegraphics[width=\linewidth]{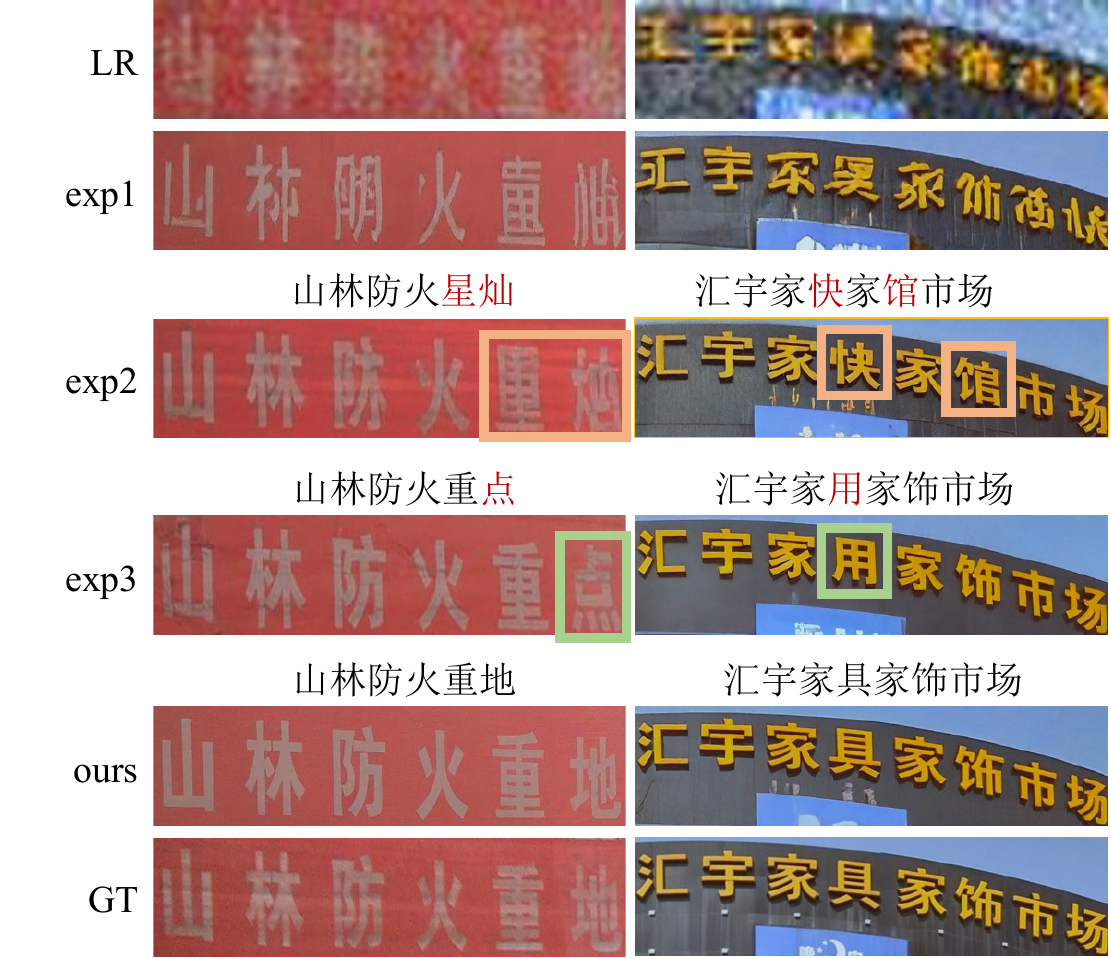}
	\caption{Ablation study to validate the effectiveness of initial text recognition (TR), TDM, and MoM.
		The text sequences above each image super-resolution result
		are the recognized text characters used for
		image super-resolution and the characters in \textcolor{red}{red} are the
		mistakenly estimated ones
		which will lead the text restoration inaccurate in the \textcolor{orange}{orange} and \textcolor{green}{green} bounding boxes.
		For the detailed settings of different methods, please see Sec.~\ref{sec:ablation}.
	}
	\label{fig:ablation}
\end{figure}

\subsection{Ablation Study} \label{sec:ablation}

In this subsection, we validate the effectiveness of different components in the proposed method and the comparison is shown in Table~\ref{tab:addlabel} and Figure~\ref{fig:ablation}.
For `exp1', it only contains IDM conditioned on the LR image.
As shown in the second row of Figure~\ref{fig:ablation}, the results look like Chinese characters.
However, the generated characters actually do not exist in the Chinese alphabet because the text prior is not considered during the diffusion process.
In `exp2' which is the baseline model in Sec.~\ref{sec:IDM}, it uses the text recognition (TR) method \cite{chen2021scene} to predict text sequence and provide text prior to IDM.
As a result, `exp2' can keep the text fidelity better than `exp1' according to the third row of Figure~\ref{fig:ablation}.
Whereas, the results are still unsatisfactory when the degradation is too severe and the text character recognition is inaccurate as seen in the \textcolor{orange}{orange} bounding boxes.
`exp3' uses TDM, whose initial state is provided by TR~\cite{chen2021scene}, to predict text sequence and provide text prior to IDM.
With the strong text sequence distribution modeling ability from TDM by diffusion, `exp3' can recognize text more accurately than `exp2' as shown in the fourth row of Figure~\ref{fig:ablation}.
But the text characters in the \textcolor{green}{green} bounding boxes are still incorrect.
This is because TDM in `exp3' does not utilize the higher-quality image information provided by IDM to recognize more accurate text sequence during the diffusion process.
Our method contains MoM module which can provide better text prior for IDM and better image prior for TDM in the diffusion steps.
In this way, TDM in our method can correct the mistakenly estimated text sequence with higher-quality image information from IDM.
At the same time, IDM can restore text image with higher fidelity which is shown in the fifth row of Figure~\ref{fig:ablation}.
Similarly, Table~\ref{tab:addlabel} shows that the proposed method can achieve consistently better performance with more components considered which demonstrates the effectiveness of TR, TDM, and MoM.

%% file: sec/5_conclusion.tex
\section{Conclusion}
In this paper, we propose to use the diffusion model to solve the blind text image super-resolution problem.
As diffusion has a strong ability to model distribution and generate data, the proposed IDM can restore realistic HR text images.
At the same time, we also apply another diffusion model (TDM) to model the distribution of text sequence and provide text prior to IDM.
In this way, IDM can also generate text images with high text fidelity.
At last, we propose MoM to make these two diffusion models appropriately cooperate with each other during the diffusion process.
Extensive experiments on synthetic and real-world datasets demonstrate our method can perform better than existing arts based on style realness and text fidelity simultaneously. 